\documentclass[letterpaper, 10 pt, conference]{ieeeconf} 
\usepackage{amsmath,amsfonts}
\usepackage{algorithmic}
\usepackage{algorithm}
\usepackage{array}
\usepackage[caption=false,font=normalsize,labelfont=sf,textfont=sf]{subfig}
\usepackage{textcomp}
\usepackage{url}
\usepackage{verbatim}
\usepackage{graphicx}
\usepackage{cite}
\usepackage{nomencl}
\usepackage{ntheorem}   
\usepackage{lipsum}     
\usepackage{graphicx}

\usepackage{xcolor}
\usepackage{hyperref}

\hypersetup{
	colorlinks=true,
	linkcolor=black,
	filecolor=black, 
	urlcolor=black,
	citecolor=black
}

\usepackage{bm}

\DeclareRobustCommand{\uvec}[1]{{%
		\ifcsname uvec#1\endcsname
		\csname uvec#1\endcsname
		\else
		\bm{\mathbf{#1}}%
		\fi
}}

\IEEEoverridecommandlockouts                              

\usepackage{graphicx} 

\title{\LARGE \bf
SPiralRoll: A Novel Adjustable-Stiffness Underactuated 3-DoF Joint with Torsion Springs for Rolling Robots
}

\author{Louis Keith, George Ripper, and Seyed Amir Tafrishi
\thanks{Louis Keith and Seyed Amir Tafrishi are with Geometric Mechanics and Mechatronics in Robotics (gm$^2$R) Lab, the School of Engineering, Cardiff University, Cardiff, CF24 3AA, United Kingdom ( e-mail:
        {\tt\small KeithL@cardiff.ac.uk} and {\tt\small Tafrishisa@cardiff.ac.uk})}
\thanks{Seyed Amir Tafrishi is the corresponding author of this study (phone: +44 29208 76176, e-mail: {\tt\small Tafrishisa@cardiff.ac.uk}).}
\thanks{*This work was supported by the Royal Society research grant under Grant \text{RGS\textbackslash R2\textbackslash 242234}.}}

\renewcommand{\baselinestretch}{.91}
\begin{document}

\maketitle
\thispagestyle{empty}
\pagestyle{empty}


\begin{abstract}
Compliant mechanisms are important in robotics because they can improve adaptability, safety, and energy efficiency while reducing hardware complexity. This paper presents SPiralRoll, a novel torsion-spring-based underactuated compliant mechanism for rolling robots and compliant robotic actuation. The mechanism uses arc-distributed elastic members and two motor inputs to realize three physically observable output motions: rotational motion, radial expansion/contraction, and axial spin induced by nonlinear compliant deformation. Two configurations, namely full-arc and single-arc designs, are developed and experimentally evaluated. Beyond benchtop validation, the mechanism is integrated into a spherical rolling robot, where proof-of-concept experiments demonstrate forward rolling and turning. The results show that the full-arc design provides better structural support and smoother deformation, whereas the single-arc design yields larger deformation and stronger inertial excitation, making it more suitable for pendulum-driven rolling locomotion. Overall, SPiralRoll provides a low-cost, compact, and fully 3D-printable solution for underactuated compliant rolling robots and adaptive robotic joints.
\end{abstract}
\section{Introduction}
Compliant actuation has become increasingly important in robotics because it can improve adaptability, safety, and energy efficiency while reducing mechanical complexity \cite{manti2016stiffening,vanderborght2013variable,wolf2015variable}. These properties are valuable in both manipulators and mobile robots, where elastic deformation can absorb disturbances, adapt morphology, and support dynamic motion generation \cite{howard2013transferring,zhong2021toward}. However, many existing compliant actuator designs remain bulky or mechanically complex, limiting their integration into compact robotic platforms \cite{mathews2022design,bons2023energy}, particularly in rolling robot applications. 

Underactuated robotics offers an attractive alternative by exploiting passive dynamics and structural intelligence to realize useful motion with fewer actuators than output degrees of freedom \cite{he2019underactuated,liu2020survey}. This is especially relevant in compact systems, where reducing actuator count can save space and weight while still enabling rich motion behaviour \cite{pustina2022feedback,xu2024tracked}. Nevertheless, achieving multiple useful output motions with a simple compliant underactuated mechanism remains challenging, particularly when the structure must also support external loads and remain easy to fabricate.

Rolling and spherical robots form a particularly relevant application class for such mechanisms. These robots offer enclosed actuation, protected hardware, and simple external morphology, making them attractive for inspection, exploration, and cluttered environments \cite{armour2006rolling,diouf2024spherical,tafrishi2019design}. Their locomotion is commonly generated through internal mass redistribution, pendulum excitation, or reaction-based actuation \cite{liu2012motion,liu2024monorollbot,tafrishi2019design}. However, most existing rolling-robot actuation strategies focus primarily on locomotion generation and do not explicitly combine internal compliant deformation, geometry-dependent stiffness, and multi-mode joint behaviour within a compact mechanism.

 \begin{figure}[t!]
    \centering
\includegraphics[width=1.02\linewidth]{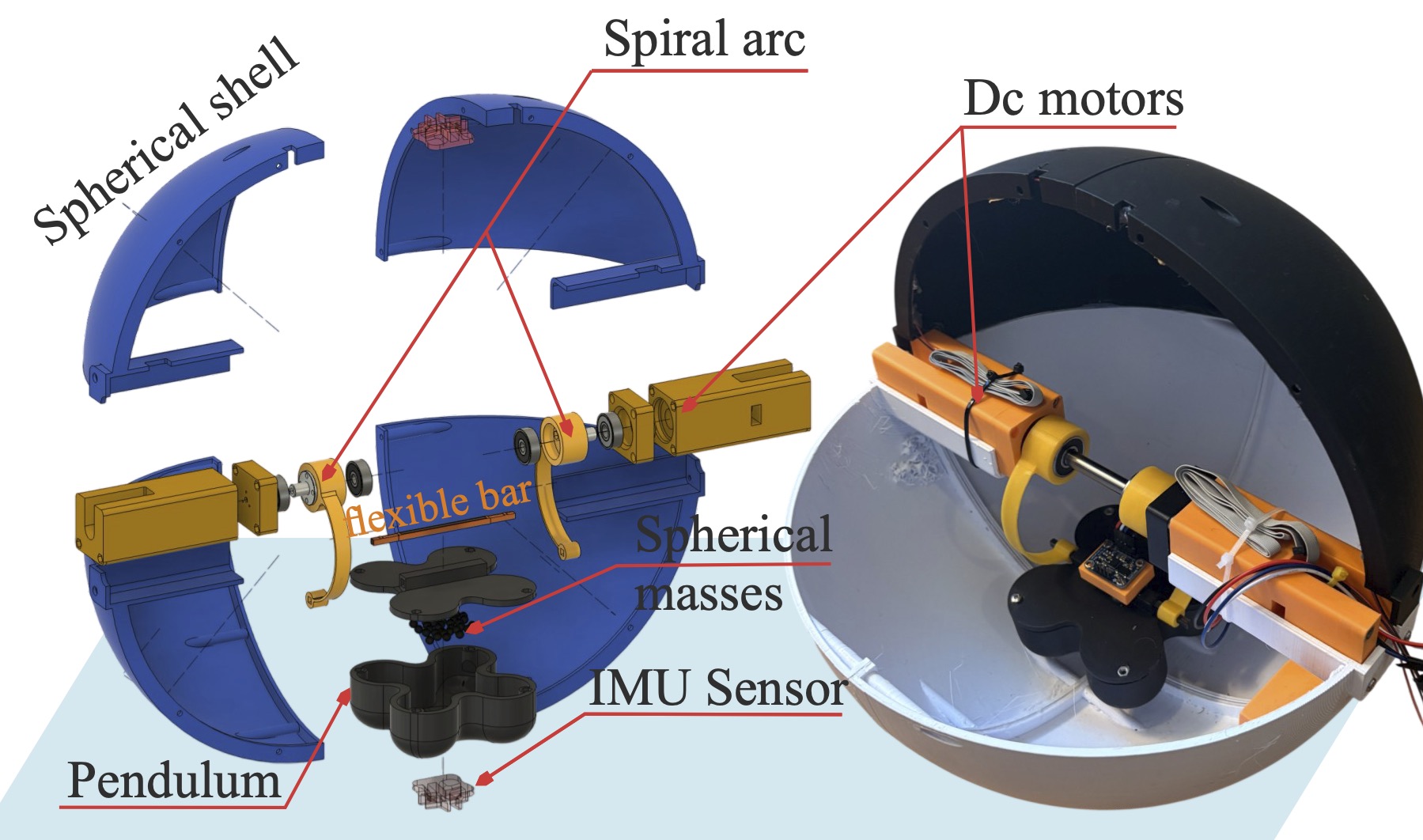}
    \caption{SPiralRoll integrated into the spherical rolling robot platform used for proof-of-concept forward rolling and turning experiments.}
    \label{fig:Spiral_Sphere}
\end{figure}

Spring-based joint designs provide a promising foundation for this goal. Spiral-spring and compliant-joint actuators can provide compact elastic behaviour, energy storage, and safer interaction \cite{kim2013compliant,10386077}. Related underactuated compliant modules and rolling platforms have also shown how geometric compliance or internal actuation can support radial deformation and spherical locomotion \cite{ugc2024,liu2024monorollbot,wang2024rollbot}. However, these studies typically emphasize either compliant joint actuation, geometric module deformation, or rolling locomotion separately. In contrast, SPiralRoll is developed as an arc-spring-based underactuated mechanism that combines rotational motion, radial expansion/contraction, and axial spin within one structure, while also being integrated into a spherical rolling robot.

Motivated by these observations, this paper introduces \emph{SPiralRoll}, a novel torsion-spring-based underactuated compliant mechanism for rolling robots, as shown in Fig.~\ref{fig:Spiral_Sphere}. The proposed mechanism distributes compliance through arc-shaped torsion-spring members arranged between two motor-driven hubs and an output bar, and is designed to realize three physically observable output motions at the load end: \emph{(i)} rotational motion, \emph{(ii)} radial expansion/contraction, and \emph{(iii)} axial spin caused by nonlinear elastic deformation. These three output motions are achieved using only two motor inputs, making the system underactuated while still capable of rich compliant behaviour. In this paper, the term 3-DoF refers to three physically observable output motions of the compliant mechanism, rather than three independently actuated kinematic coordinates in the conventional robotics sense. To study this concept, two prototype configurations are investigated: a \emph{full-arc} configuration emphasizing structural support and distributed compliance, and a \emph{single-arc} configuration emphasizing stronger deformation and pendulum-like excitation. In addition to benchtop validation, the mechanism is integrated into a spherical rolling robot to assess its applicability to forward rolling and turning. The main contributions of this paper are therefore: 1) the design of a compact fully 3D-printable torsion-spring-based underactuated compliant mechanism, presented in Section~\ref{Sec:MechanismDesign}; 2) experimental demonstration of rotational, radial, and spin output motions, reported in Section~\ref{Sec:MotionAnalysis}; 3) comparative evaluation of the full-arc and single-arc configurations in terms of deformation range, support, and dynamic response, discussed in Section~\ref{Sec:MotionAnalysis}; and 4) proof-of-concept spherical rolling robot experiments demonstrating practical rolling applicability, presented in Section~\ref{Sec:MotionAnalysis}.
\section{Mechanism Design and Model}
\label{Sec:MechanismDesign}

This section presents the SPiralRoll mechanism, its two prototype configurations, and the spherical rolling robot implementation used to assess practical applicability.

 \begin{figure}[t!]
    \centering
\includegraphics[width=1.05\linewidth]{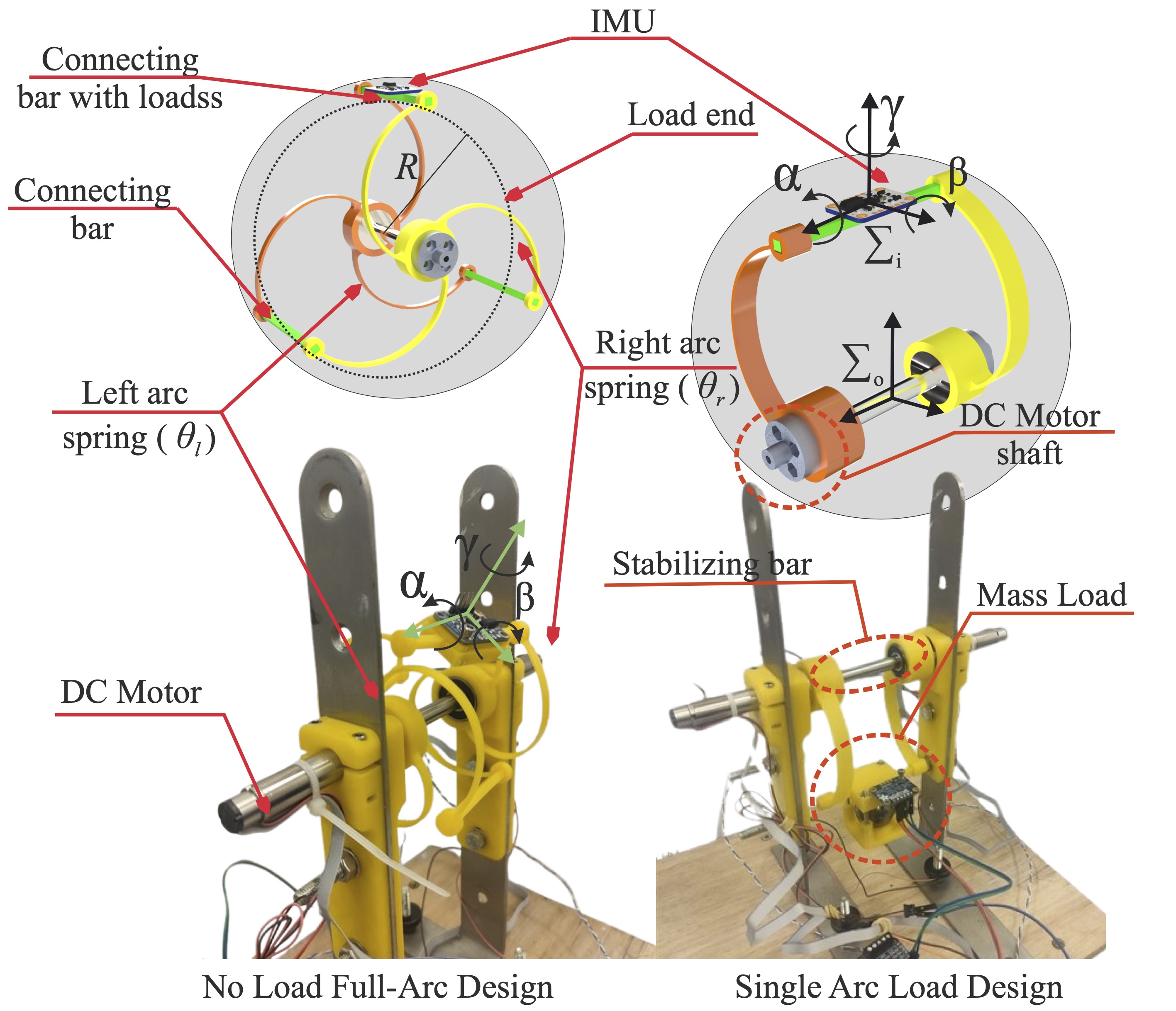}
    \caption{SPiralRoll mechanism and its two prototype configurations. The full-arc design emphasizes distributed compliance and structural support, whereas the single-arc design is intended for stronger deformation and pendulum-like excitation.}
    \label{fig:straightbarjoint}
\end{figure}

The proposed mechanism is inspired by earlier spiral-spring-based joint concepts \cite{tafrishi2019design}, while addressing a common limitation of conventional compact spiral layouts: deformation is typically concentrated between the motor shaft and the outer spring coupling, which can reduce load support and restrict integration into mobile systems. In many existing designs, compliance is mainly exploited for rotation about a dominant axis. By contrast, SPiralRoll distributes elasticity through arc-shaped members so that compliance also contributes to geometric reconfiguration and momentum transfer, which is particularly useful in underactuated rolling mechanisms.

The output behavior of the mechanism is characterized by three physically distinct and experimentally observable motion modes. The first is the rotational motion of the output bar when the two motors rotate in the same direction with similar velocity. The second is the radial expansion or contraction of the compliant arc structure when the motors are driven differentially, leading to a measurable change in the effective radius $R$. The third is the axial spin or twisting motion of the output bar caused by nonlinear elastic deformation and asymmetric stress distribution within the arc-spring members. These three output motions are achieved using only two motors, and the mechanism is therefore underactuated. Accordingly, the mechanism is best interpreted as exhibiting three measurable output motions generated by two actuators, with the axial spin appearing as an emergent deformation mode under asymmetric loading.

As shown in Fig.~\ref{fig:straightbarjoint}, the mechanism consists of two motor-driven arc-spring branches connected at the load end by an output bar. The left and right motor rotations are denoted by $\theta_l$ and $\theta_r$, respectively. Coordinated or differential actuation deforms the compliant structure and generates the three output motions described above. The output bar can carry a payload, pendulum mass, or inertial measurement unit (IMU), allowing the internal deformation to be directly transferred to the attached load.

Two physical configurations are investigated. The \emph{full-arc} design uses three arc springs on each side to improve load support, distribute deformation more evenly, and preserve structural stability. The \emph{single-arc} design uses one thicker arc on each side to generate stronger deformation and more pronounced pendulum-like excitation, making it more suitable for rolling applications. The main geometric and stiffness-related properties of both designs are summarized in Table~\ref{table:dimensions}. The geometric parameters in Table~\ref{table:dimensions} were selected through iterative prototyping subject to practical constraints, including motor torque capability, printable geometry limits, desired deformation range, and integration requirements for the spherical robot platform.

\begin{table}[t!]
\centering
\begin{tabular}{ l l l }
\hline
\hline
\textbf{Dimension} & \textbf{Full-Arc} & \textbf{Single-Arc} \\
\hline
Arc spring thickness, $h_s$ (mm) & 4 & 10 \\
\hline
Arc spring width, $w_s$ (mm) & 1.5 & 1.5 \\
\hline
Hub diameter, $d_h$ (mm) & 10 & 10 \\
\hline
Arc length, $d_c$ (mm) & 88.62 & 88.62 \\
\hline
Initial arc radius, $R$ (mm) & 50 & 50 \\
\hline
Connecting bar length (mm) & 60 & 60 \\
\hline
Stabilizing bar length (mm) & 58 & 58 \\
\hline
Stiffness constant of each arc, $K_c$ (N/m) & 0.451 & 7.053 \\
\hline
Total side stiffness constant, $K_{ct}$ (N/m) & 1.353 & 7.053 \\
\hline
\hline
\end{tabular}
\caption{Dimensions of the two SPiralRoll actuator configurations.}
\label{table:dimensions}
\end{table}

The stabilizing bar improves structural integrity and helps maintain controlled deformation during motion. As a result, the full-arc design favors smoother deformation and better load support, whereas the single-arc design concentrates compliance and produces larger deformation amplitudes. This distinction is important for rolling robots, where stronger internal deformation can improve pendulum-driven excitation \cite{armour2006rolling}.

Based on this structure, the three DoFs emerge in a physically intuitive way. The first DoF corresponds to free rotation of the load when both motors operate with approximately similar velocity ($\dot{\theta}_l \approx \dot{\theta}_r$). The second DoF corresponds to contraction and expansion of the mechanism, observed as a change in radius $R$, when the motor angles differ significantly ($\theta_r \neq \theta_l$). The third DoF arises from the nonlinear flexibility of the arc members, which allows the output bar to twist or spin axially under asymmetric deformation. Since two actuators generate three physically observable output motions, the mechanism is underactuated by design.

In this mechanism, Poly Lactic Acid (PLA) is used to fabricate the 3D-printed double-sided arc spring, providing sufficient stiffness while remaining lightweight and easy to manufacture. PLA was selected for this proof-of-concept study because it provides a practical compromise between stiffness, low cost, ease of fabrication, and repeatability in rapid prototyping \cite{10386077}. The stiffness constant of each arc spring assuming linear deflection over angle is determined using the following formula \cite{kim2013compliant}:
\begin{equation}
K_{c} = \frac{\eta w_s h_s^3}{ 12 d_c },
\label{Eq:Stifnessconstant}
\end{equation}
where $\eta$, $w_s$, $h_s$, and $d_c$ represent Young's modulus, the width of the spring, the arc spring thickness, and the initial arc radius of the spiral spring, respectively. Note that the stiffness value of $K_{c}$ has been validated through deflection and torque experiments. Based on the material type (PLA), the Young's modulus $\eta$ is 5 GPa. This allows us to compute the approximate stiffness profile, as presented in Table \ref{table:dimensions}. It is evident that the single arc design, due to its increased width in a pendulum configuration, exhibits a higher stiffness profile. However, it is important to note that different stiffness values exist for each side of the motor connection, as there are three arcs per side, resulting in a combined stiffness constant of 1.353 N/m.

To analyze the rotational motion and degrees of freedom of the load side in the designed mechanism, we utilize an IMU sensor mounted at the load end of the underactuated compliant actuator. The IMU measures its orientation relative to the inertial frame \( \Sigma_o \) and provides the Euler angles \( (\gamma, \beta, \alpha) \), which describe the rotational state of the actuator’s load. The transformation from the local frame \( \Sigma_i \), attached to the actuator, to the inertial frame \( \Sigma_o \) is defined by the following rotation matrix:  
\begin{align}
 & \mathbf{R}(\gamma,\beta,\alpha) = \mathbf{R}_z(\gamma) \mathbf{R}_y(\beta) \mathbf{R}_x(\alpha) \nonumber\\  
 & = \begin{bmatrix}
C_\alpha C_\beta & C_\alpha S_\beta S_\gamma - S_\alpha C_\gamma & C_\alpha S_\beta C_\gamma + S_\alpha S_\gamma \\
S_\alpha C_\beta & S_\alpha S_\beta S_\gamma + C_\alpha C_\gamma & S_\alpha S_\beta C_\gamma - C_\alpha S_\gamma \\
-S_\beta & C_\beta S_\gamma & C_\beta C_\gamma 
\end{bmatrix},
\end{align}
where \( C \) and \( S \) denote the cosine and sine functions, respectively.  

Furthermore, to evaluate the total force \( F_n \) generated by the rotating load, we utilize the filtered IMU acceleration \( \mathbf{a}(t) \), processed through a zero-low pass filter. The force $F_n$ is approximated as  
\begin{equation}
F_n \approx m_b \parallel \mathbf{R}(t) \; \mathbf{a}(t) - \mathbf{R}(t_0) \; \mathbf{a}_g \parallel,  
\end{equation}
where \( \mathbf{a}_g \) represents the initial acceleration measurement at time \( t_0 \). This formulation allows us to estimate the dynamic response of the actuator’s load based on IMU readings.  

An alternative end-effector was developed to exploit the axial rotation of the mechanism more effectively in spherical rolling operation. During expansion and contraction, rotation about the $\gamma$ axis produces a center-of-mass (CoM) shift, which can induce lateral acceleration and contribute to turning motion. The proposed end-effector is a hollow PLA X-shaped body loaded with ball-bearing masses to increase the CoM shift. A granular mass set was also considered as an alternative to a single rigid spherical mass, since internal relative motion within the granular medium can provide damping and stabilization effects \cite{granularrobotics,wang2024rollbot}. The redistributed mass is intended to move toward the corners of the X-shaped body during rotation, thereby amplifying the CoM shift and improving turning authority. The main geometric properties of the spherical robot implementation are summarized in Table~\ref{table:sphere_dim}. The parameters in Table~\ref{table:sphere_dim} were selected to ensure mechanical integration within the spherical shell, sufficient pendulum travel, and compatibility with the motor housings, bearings, and internal end-effector geometry.

\begin{table}[t!]
\centering
\resizebox{\linewidth}{!}{
\begin{tabular}{ l l }
\hline
\hline
\textbf{Dimension} & \textbf{Full Sphere System} \\
\hline
Arc spring thickness, $h_s$ (mm) & 10 \\
\hline
Arc spring width, $w_s$ (mm) & 1.5 \\
\hline
Arc length, $d_c$ (mm) & 88.62 \\
\hline
Initial arc radius, $R_s$ (mm) & 50 \\
\hline
Stabilizing bar length (mm) & 55 \\
\hline
Connecting bar length (mm) & 60 \\
\hline
Sphere radius, $R_s$ (mm) & 145 \\
\hline 
Pendulum width, $w_{EE}$ (mm) & 94 \\
\hline
Pendulum height, $h_{EE}$ (mm) & 47 \\
\hline
Stiffness constant $K_c$ (N/m) & 7.053 \\
\hline
Total side stiffness constant $K_{ct}$ (N/m) & 7.053 \\
\hline
\hline
\end{tabular}
}
\caption{Dimensions of the spherical rolling robot system.}
\label{table:sphere_dim}
\end{table}

To evaluate the practical applicability of the proposed mechanism, the single-arc SPiralRoll design was integrated into a PLA spherical shell platform, as shown in Fig.~\ref{fig:Spiral_Sphere}. The shell has a radius of 145 mm, and the mechanism is mounted between two internal PLA housings that also support the motors and bearings. In this arrangement, the compliant joint acts as an internal pendulum-like actuator, while the modified end-effector enables controllable CoM redistribution for both forward rolling and turning. The IMU is mounted centrally on the top cap of the end-effector so that internal pendulum motion can be related directly to the measured behavior of the sphere. The complete system is operated using a MATLAB script, in which time-based motor input profiles are applied to generate different motion primitives. This spherical implementation is not intended as a fully optimized rolling robot, but rather as a focused proof-of-concept demonstrating that the proposed SPiralRoll mechanism can function as a compact internal actuation system for rolling locomotion.
\section{Motion Analysis}
\label{Sec:MotionAnalysis}

In this section, we evaluate the motion capabilities of the proposed actuator and examine how its compliance influences the three observable degrees of freedom. We also assess its suitability as a compliant pendulum-like actuator for rolling robots, where internal deformation and momentum transfer are directly linked to locomotion performance.

The experimental setup employs 6~V Maxon DC motors for actuation and an Adafruit BNO055 absolute orientation sensor for real-time motion tracking. The motors are driven through an Arduino Uno and motor driver shield, while MATLAB is used for command generation and data acquisition. To study the response under different actuation levels, step-input tests are performed at no power, half power, and full power. For experiments involving a load, a spherical mass of $m_b = 0.15$~kg is attached to the output bar. These tests are used to characterize the mechanism's rotational, radial, and twisting responses, as well as its dynamic suitability for mobile robotic applications.

\begin{figure}[t!]
    \centering
    \includegraphics[width=3.4 in]{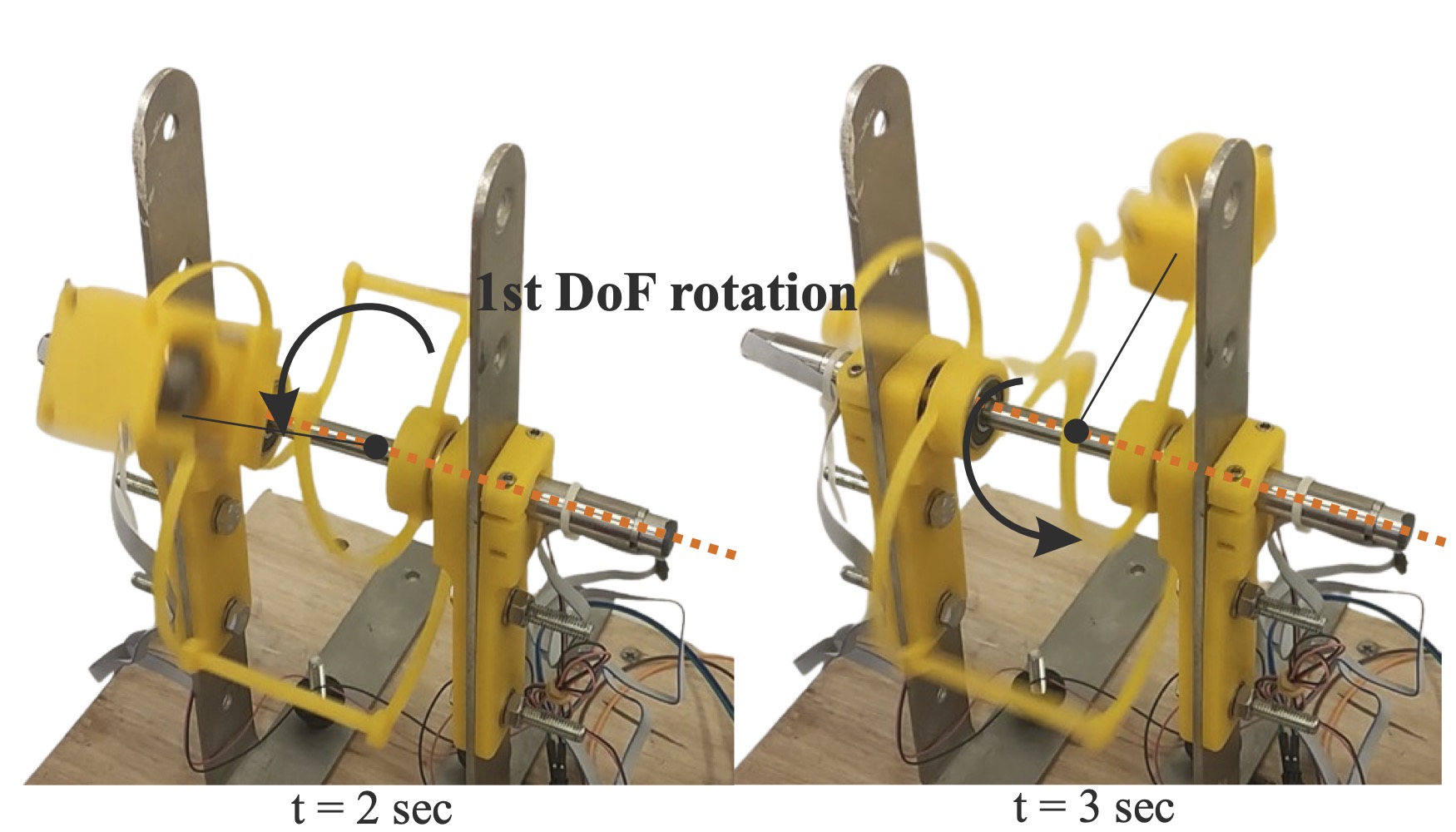}
    \includegraphics[width=2.1 in]{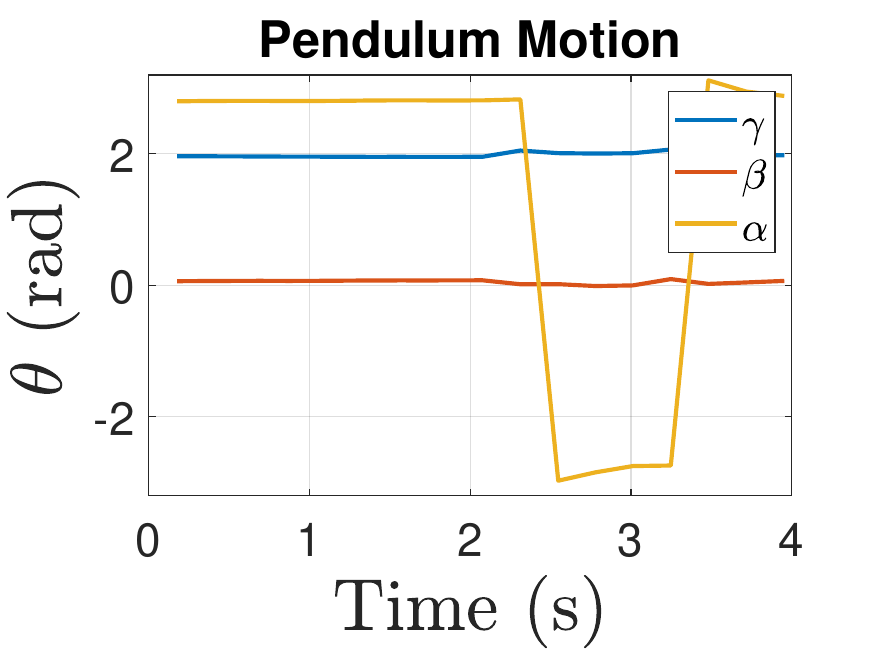}
    \caption{Validation of the 1st DoF using the full-arc configuration with an attached mass. Coordinated motor actuation produces rotational motion of the output bar about the $\alpha$ axis.}
    \label{fig:Basicrotaion}
\end{figure}

\subsection{Degree of Freedom in Full-Arc}

We first evaluate the full-arc configuration, which uses three compliant arc springs on each side. The initial test focuses on the first degree of freedom, namely rotational motion of the outer load. This motion is generated when both motors rotate with similar velocity and in the same direction. Figure~\ref{fig:Basicrotaion} shows a representative example. The results indicate that the full-arc mechanism preserves its overall geometry while producing a clear rotational response about the $\alpha$ axis.

\begin{figure}[t!]
    \centering
    \includegraphics[width=1.2 in]{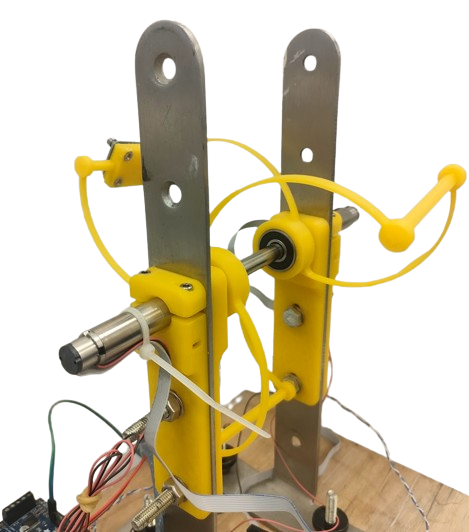}
    \includegraphics[width=1.2 in]{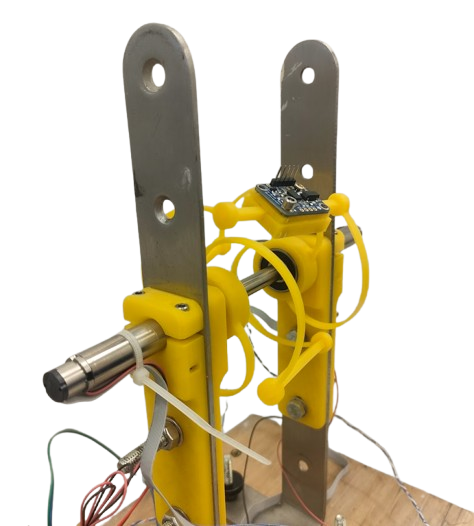}
    \includegraphics[width=1.6 in]{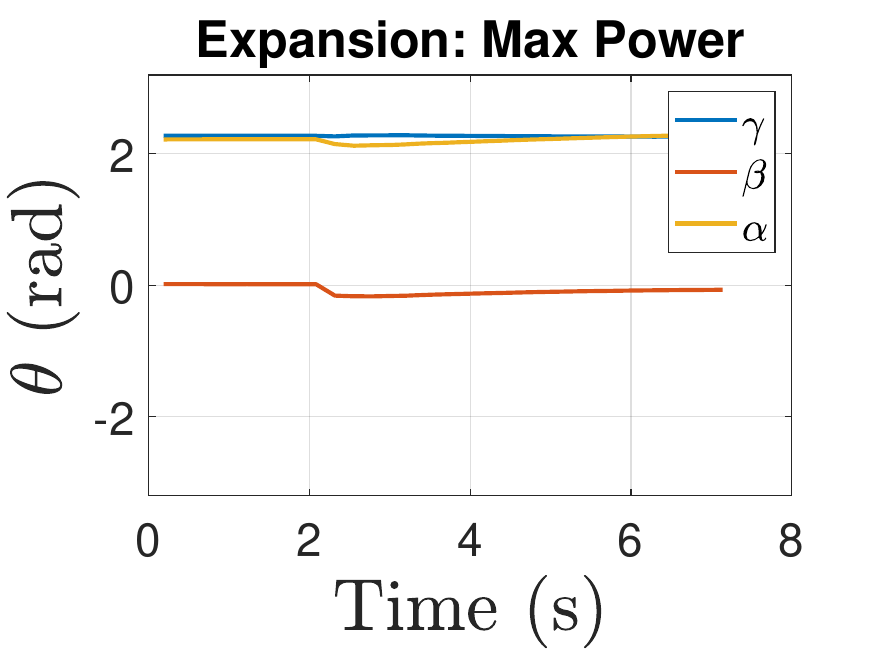}
    \includegraphics[width=1.6 in]{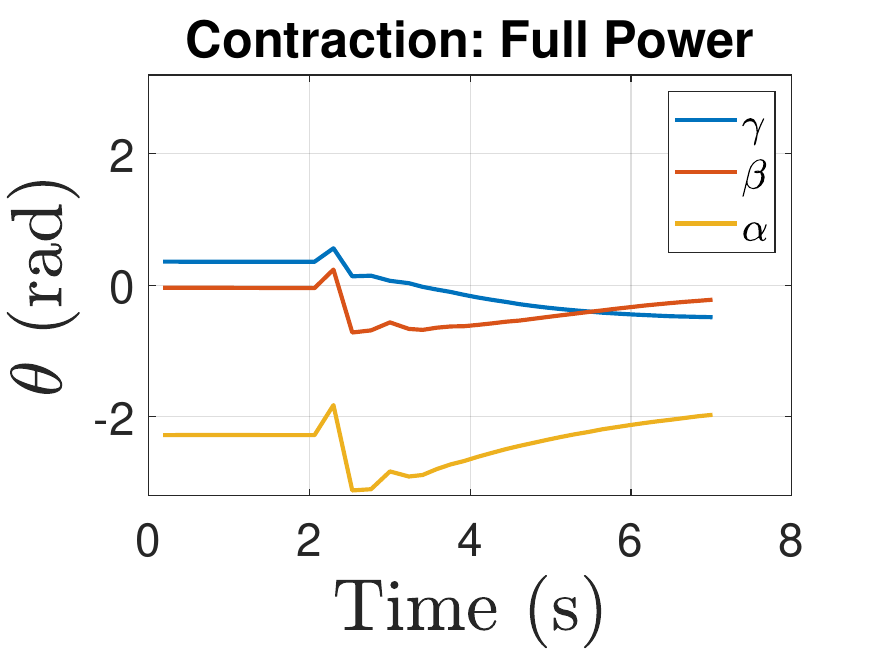}
    \caption{Validation of the 2nd DoF for the full-arc configuration. Differential motor actuation causes radial expansion and contraction, corresponding to a measurable change in effective radius $R$.}
    \label{fig:expansionTriple2}
\end{figure}

The second degree of freedom is examined by driving the two motors in opposite directions. In this case, the mechanism undergoes contraction or expansion, producing a clear change in the effective radius $R$, as shown in Fig.~\ref{fig:expansionTriple2}. The maximum deformation capability is summarized in Table~\ref{table:diameters}, where the mechanism achieves approximately 50\% variation relative to the reference diameter of $2R = 148$~mm. This quantity is measured using visual analysis, since the IMU alone cannot capture axial displacement directly.

\begin{figure}[t!]
    \centering
    \includegraphics[width=1.4 in]{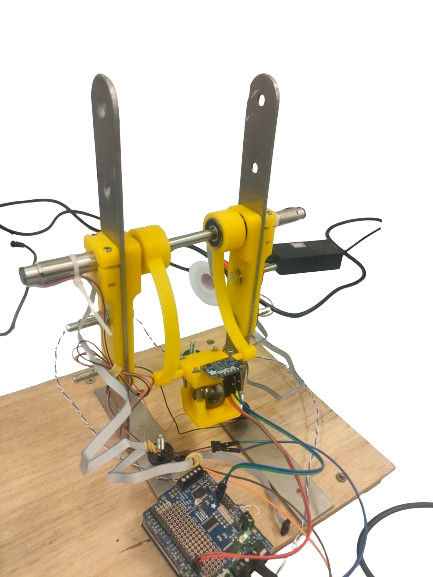}
    \includegraphics[width=1.4 in]{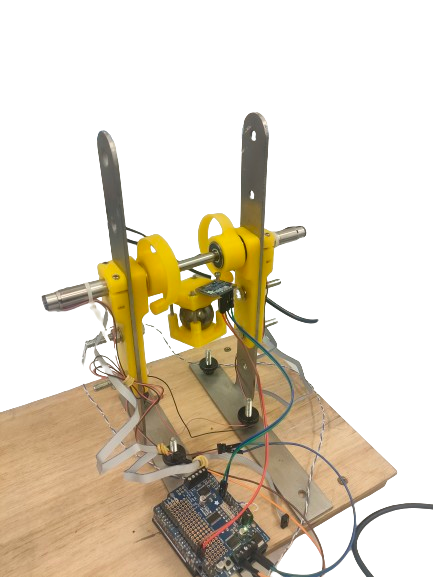}\\
    \includegraphics[width=1.5 in]{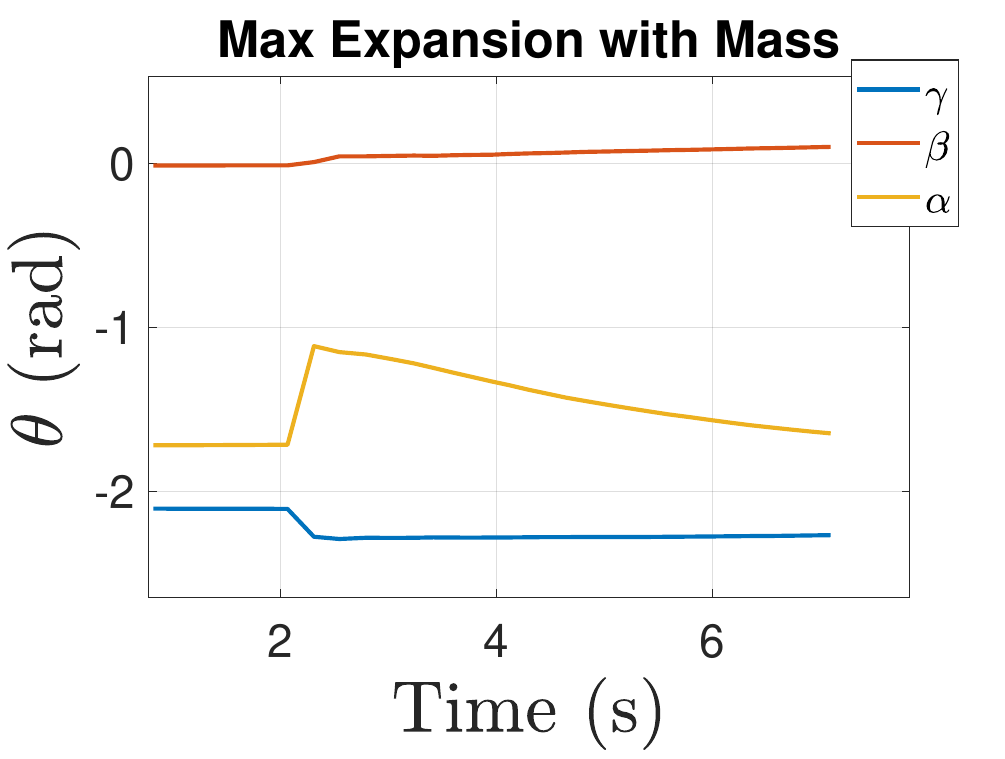}
    \includegraphics[width=1.5 in]{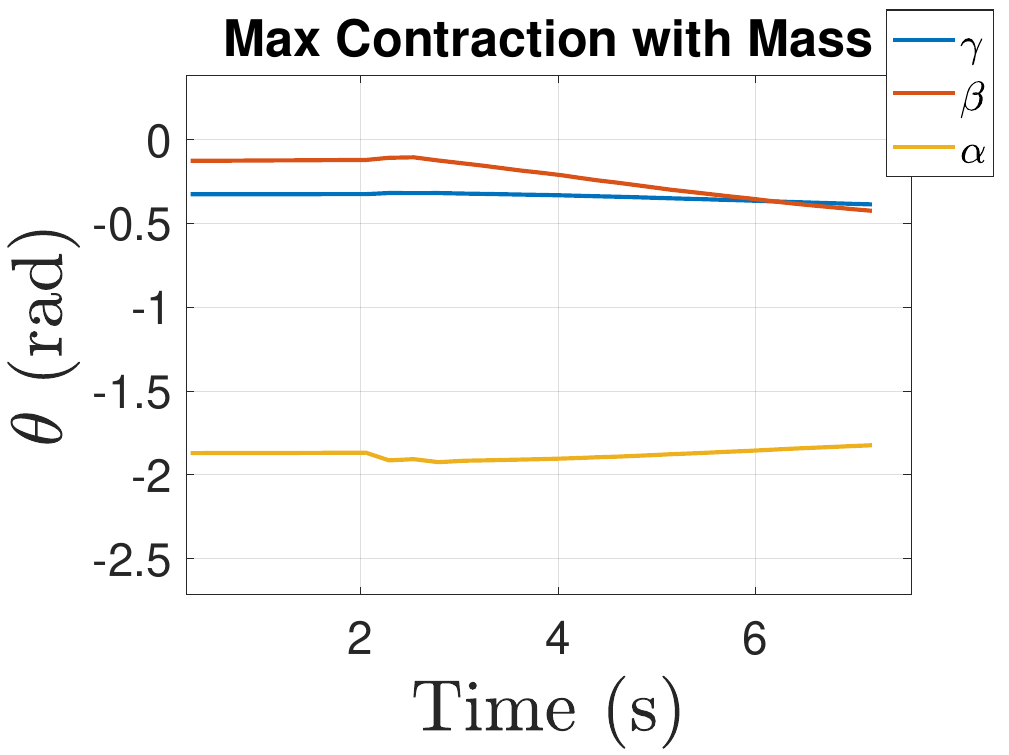}
    \caption{Loaded single-arc mechanism during expansion and contraction. The figure highlights the coupled radial deformation and axial spin response associated with the third observable output motion.}
    \label{fig:MassedSingleExpansion}
\end{figure}

The third degree of freedom emerges when the angular difference between the two motors, $\theta_l$ and $\theta_r$, exceeds a critical threshold. Under this condition, nonlinear flexibility and asymmetric stress distribution within the arc members induce axial spin, represented by the angle $\gamma$. This behavior is visible in both the full-arc and single-arc experiments, and is especially pronounced in the loaded single-arc case shown in Fig.~\ref{fig:MassedSingleExpansion}. Experimentally, the structure exhibits approximately 1~rad of $\gamma$ variation. Although a complete nonlinear model of this coupling has not yet been developed, the repeated observation of this motion confirms that axial spin is an inherent output mode of the mechanism.

\begin{table}[t!]
\centering
\begin{tabular}{l  l  l  l }
\hline
\hline
\textbf{Scenario} & \textbf{2R (mm)} & \textbf{Change (mm)} & \textbf{Percentage Change} \\
\hline
Reference & 148 & NA & NA \\
\hline
Full expansion & 160 & +12 & +8.1\% \\
\hline
Full contraction & 82 & -66 & -44.6\% \\
\hline
Half contraction & 126 & -22 & -14.9\% \\
\hline
\hline
\end{tabular}
\caption{Diameter variation of the full-arc configuration under different actuation scenarios.}
\label{table:diameters}
\end{table}

\begin{figure}[t!]
    \centering
    \includegraphics[width=1.6 in]{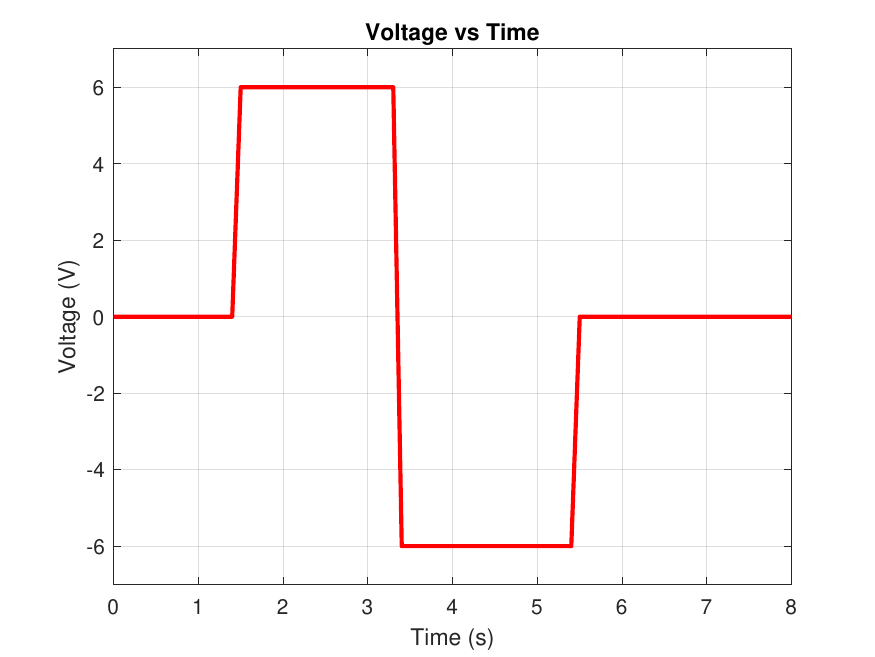}
    \includegraphics[width=1.7 in]{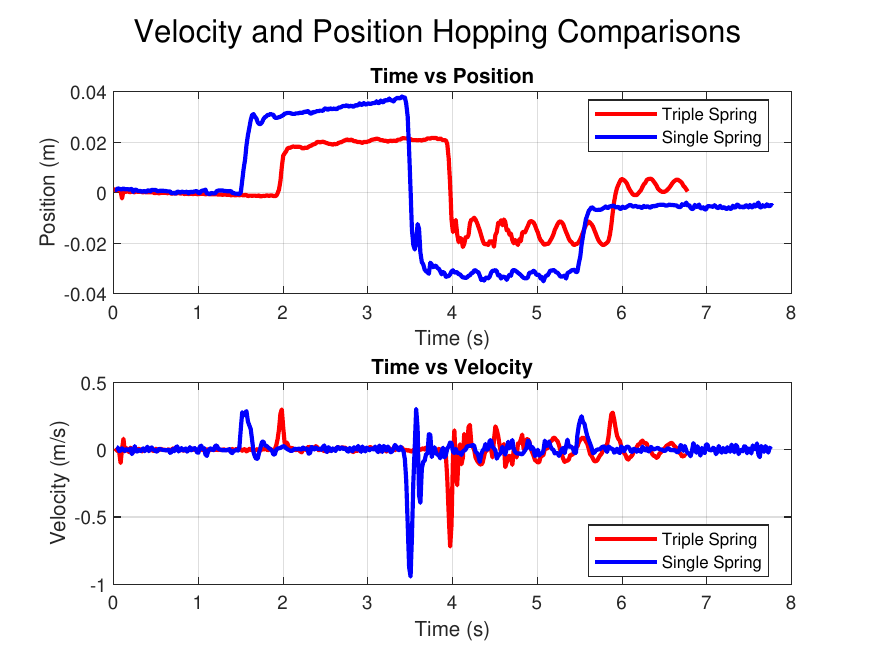}
    \caption{Dynamic-response comparison of the full-arc and single-arc configurations. Left: commanded motor input profile. Right: visually estimated position and velocity responses under hopping-like excitation, showing stronger inertial response in the single-arc case.}
    \label{fig:MotionAnalysisSingle}
\end{figure}

\subsection{Pendulum-Mode Single Arc}

We next analyze the single-arc configuration, which is intended to provide stronger deformation and pendulum-like excitation. Compared with the full-arc design, the single-arc structure transfers motor motion through a more concentrated compliant path, leading to larger deformation amplitudes and stronger inertial effects.

\begin{table}[t!]
\centering
\begin{tabular}{ l l l l}
\hline
\hline
\textbf{Scenario} & \textbf{2R (mm)} & \textbf{Change (mm)} & \textbf{Percentage Change} \\
\hline
Reference & 78 & NA & NA \\
\hline
Full expansion & 89 & +11 & +14\% \\
\hline
Half expansion & 83 & +5 & +6\% \\
\hline
Full contraction & 28 & -50 & -64\% \\
\hline
Half contraction & 68 & -10 & -13\% \\
\hline
\hline
\end{tabular}
\caption{Diameter variation of the single-arc configuration with an added mass of $m_b = 150$~g.}
\label{table:lengths150}
\end{table}

Figure~\ref{fig:MassedSingleExpansion} shows the loaded single-arc design during expansion and contraction. The mechanism exhibits slight deviation of the rotation axis while preserving a clear twisting response characterized by $\gamma$. This twisting becomes more pronounced during contraction, while the structure remains sufficiently symmetric to maintain controlled deformation. The results in Table~\ref{table:lengths150} show that the single-arc design achieves approximately 60--70\% radius variation, despite its higher stiffness constant $K_{ct}$. This makes it particularly attractive for pendulum-actuated rolling systems, where stronger deformation can improve internal momentum generation \cite{armour2006rolling}.

To compare the dynamic response of the two configurations, a controlled hopping-like input is applied to both motors, as shown in Fig.~\ref{fig:MotionAnalysisSingle}. The resulting position and velocity are estimated using visual motion tracking. The single-arc configuration produces a larger dynamic response, likely due to its higher stiffness and more concentrated deformation path. By contrast, the full-arc design provides better support and smoother motion, but exhibits lower-amplitude oscillation during convergence.

\begin{figure}[t!]
    \centering
    \includegraphics[width=0.5\linewidth]{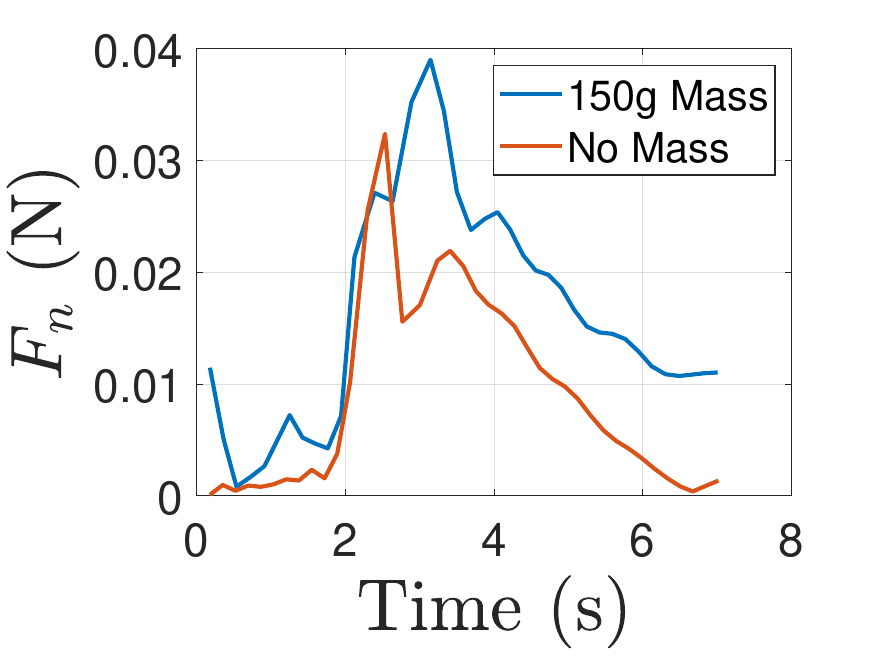}
    \caption{IMU-based dynamic response during twisting and contraction tests. Increased load produces larger inertial response, supporting the pendulum-like actuation interpretation.}
    \label{fig:ForcePlot}
\end{figure}

Figure~\ref{fig:ForcePlot} further illustrates the IMU-based dynamic response during twisting experiments. The measured response increases with added mass, indicating stronger inertial excitation. Overall, the single-arc design is more suitable for rolling applications requiring stronger internal momentum transfer, whereas the full-arc design is better suited to joint-like actuation where structural support and smoother motion are more important.

\subsection{Spherical Robot Experiment}
To evaluate the practical applicability of the mechanism, two spherical rolling experiments were performed using the single-arc SPiralRoll configuration and the X-shaped granular end-effector. The first experiment examined straight forward rolling, while the second examined turning behavior. In both cases, open-loop motor commands were used to deform the internal mechanism and redistribute the $0.06$~kg granular mass inside the end-effector. These tests are intended as proof-of-concept demonstrations rather than as a full quantitative locomotion benchmark.

\begin{figure}[t!]
    \centering
    \includegraphics[width=3.2 in, height=3.4 in]{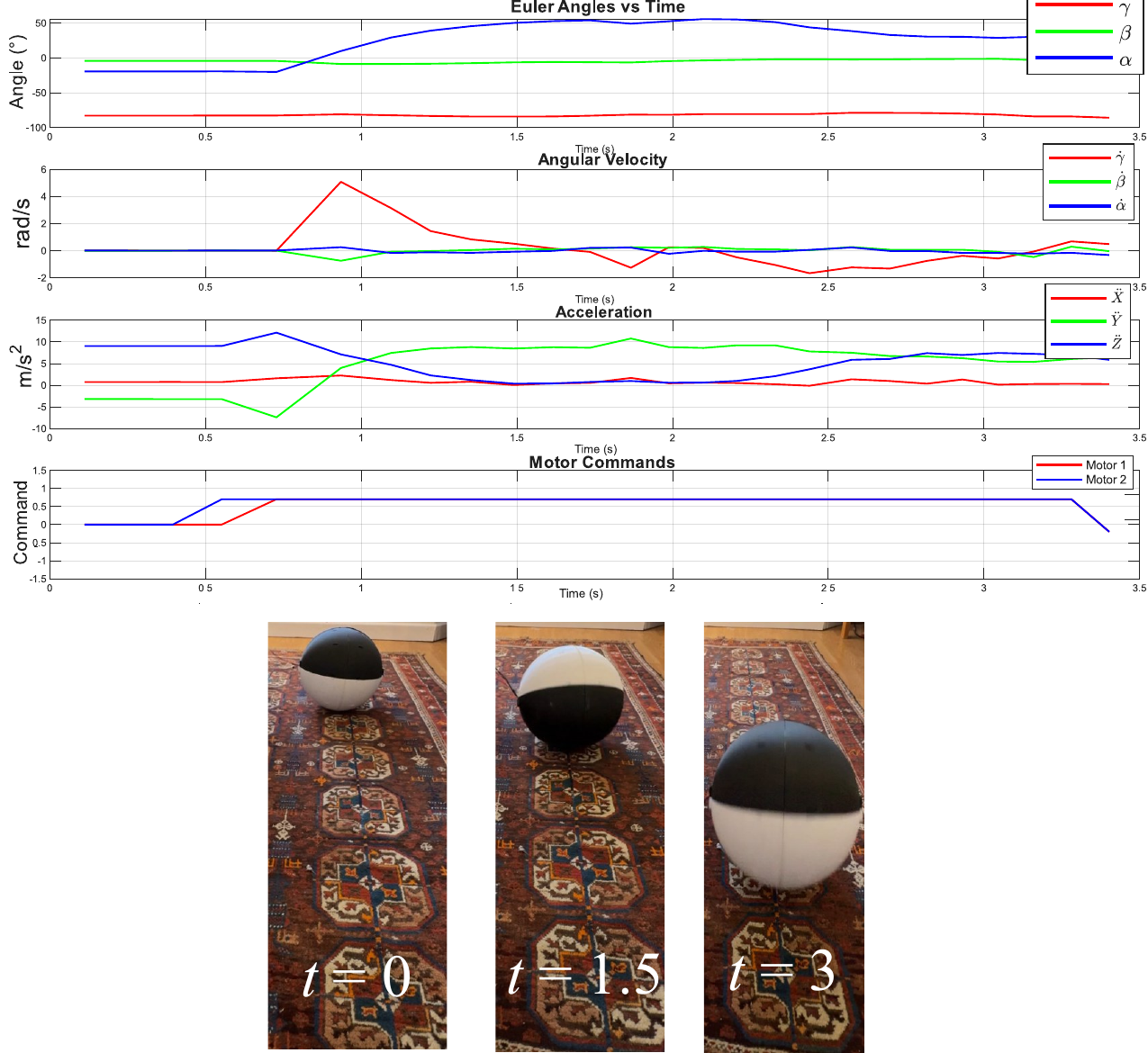}
    \caption{Spherical rolling robot experiment: forward rolling under near-symmetric actuation.}
    \label{fig:Sphere_Roll_Louise1}
\end{figure}

In the forward rolling experiment, shown in Fig.~\ref{fig:Sphere_Roll_Louise1}, the motors were supplied with approximately equal voltages, with a $0.1$~s delay applied to the left motor as an empirical phase compensation for the small asymmetry observed in the assembled spring response. This offset was introduced at the command level to improve the consistency of forward rolling in the current prototype.

\begin{figure}[t!]
    \centering
    \includegraphics[width=3.2 in, height=3.3 in]{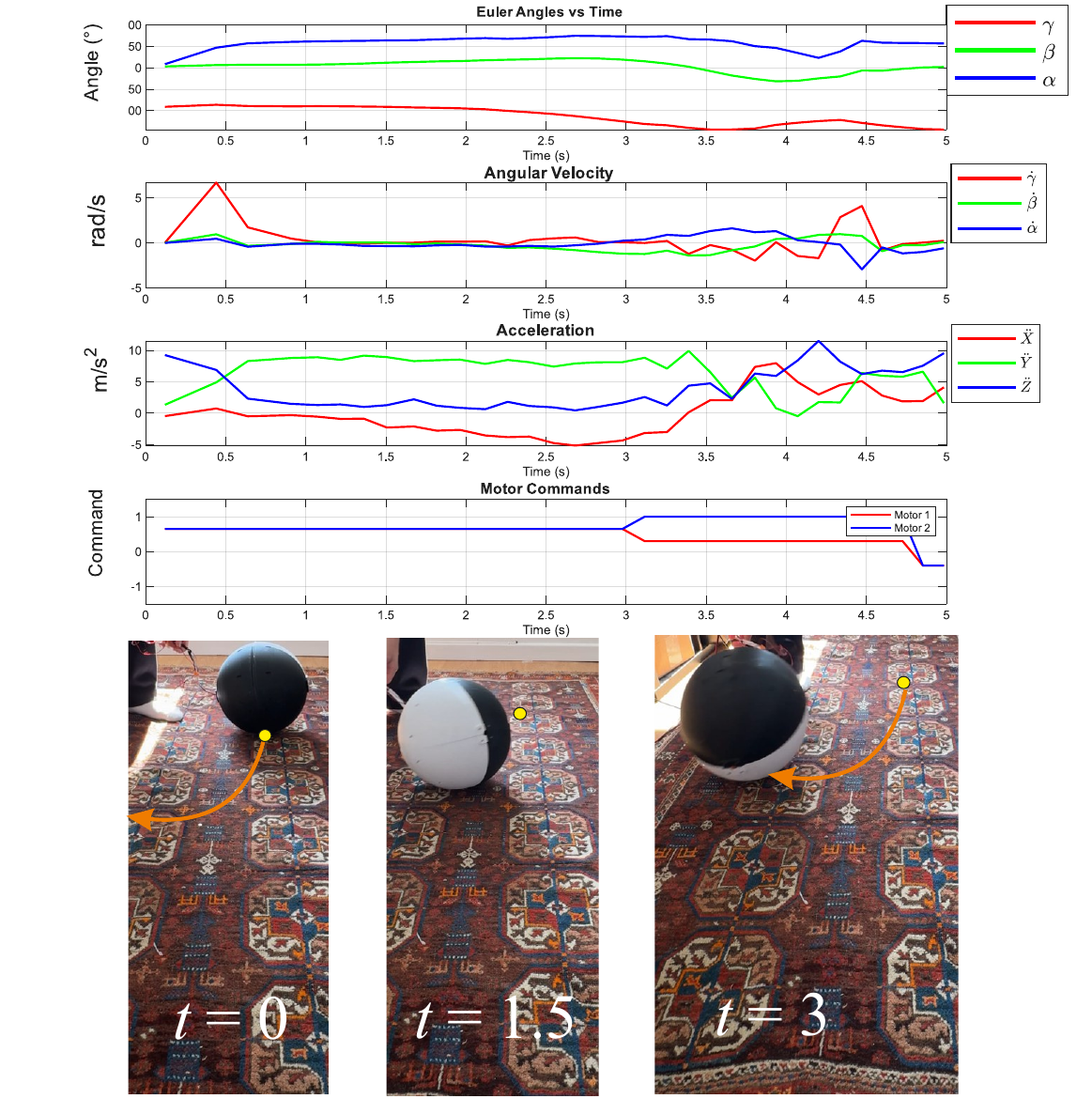}
    \caption{Spherical rolling robot experiment: right turning under asymmetric actuation.}
    \label{fig:Sphere_Roll_Louise2}
\end{figure}

In the turning experiment, shown in Fig.~\ref{fig:Sphere_Roll_Louise2}, an asymmetric motor command was applied after approximately 3~s of forward rolling. Specifically, $\theta_r$ was increased to $1.0$ while $\theta_l$ was reduced to $0.3$, producing a contracted state while the pendulum was still rotating about $\alpha$. This generated a strong response about the $\gamma$ axis, with a change of approximately $0.93$~rad ($53.3^\circ$) and a total range of approximately $1.01$~rad ($57.9^\circ$). As seen in Fig.~\ref{fig:Sphere_Roll_Louise2}, this produced a clear right-turning maneuver of approximately $53.1^\circ$.

These proof-of-concept results indicate that the proposed SPiralRoll mechanism can function as an internal actuation method for spherical rolling robots. The combination of compliant deformation and the X-shaped granular end-effector generates sufficient center-of-mass shift to achieve both forward rolling and turning. This supports the broader claim that the proposed underactuated compliant mechanism is not only effective as a benchtop joint concept, but also suitable for preliminary spherical locomotion demonstration. The present rolling experiments are intentionally limited to open-loop proof-of-concept demonstrations and are not intended as a full quantitative locomotion benchmark.

\section{Conclusion}
In this work, we introduced SPiralRoll, a novel underactuated compliant mechanism for joint actuation and rolling robots, exhibiting three physically observable output motions. The proposed design employs compliant torsion-based arc-spring structures to realize rotation, radial expansion/contraction, and axial spin using only two motor actuators. Experimental results confirmed that the mechanism can generate these motions while providing sufficient compliance, geometric adaptability, and momentum transfer for both benchtop actuation and rolling applications.

The comparison between the two configurations revealed a clear design trade-off. The single-arc configuration exhibited higher stiffness, larger deformation, and stronger inertial excitation, making it more suitable for pendulum-driven rolling locomotion. In contrast, the full-arc configuration provided better load support, more stable multi-point deformation, and smoother motion, making it more attractive for compliant joint-like actuation. The spherical robot experiments further showed that the proposed mechanism can serve as a compact internal actuation system capable of producing proof-of-concept forward rolling and turning through controlled center-of-mass shift.

Although the present study validates the mechanism experimentally, its nonlinear deformation and stiffness coupling have not yet been fully modeled. Future work will therefore focus on developing a comprehensive dynamic model, improving stiffness and torque characterization, and designing robust control strategies for more precise motion generation in rolling robot platforms.

\bibliographystyle{IEEEtran}
 \bibliography{reference}

@article{diouf2024spherical,
  title={Spherical rolling robots---Design, modeling, and control: A systematic literature review},
  author={Diouf, Aminata and Belzile, Bruno and Saad, Maarouf and St-Onge, David},
  journal={Robotics and Autonomous Systems},
  volume={176},
  pages={104657},
  year={2024},
  doi={10.1016/j.robot.2024.104657}
}

@article{pustina2022feedback,
  title={Feedback regulation of elastically decoupled underactuated soft robots},
  author={Pustina, Pietro and Della Santina, Cosimo and De Luca, Alessandro},
  journal={IEEE Robotics and Automation Letters},
  volume={7},
  number={2},
  pages={4512--4519},
  year={2022},
  publisher={IEEE}
}

@article{armour2006rolling,
  title={Rolling in nature and robotics: A review},
  author={Armour, Rhodri H and Vincent, Julian FV},
  journal={Journal of Bionic Engineering},
  volume={3},
  number={4},
  pages={195--208},
  year={2006},
  publisher={Springer}
}

@inproceedings{liu2012motion,
  title={Motion control of an underactuated spherical robot: A hierarchical sliding-mode approach with disturbance estimation},
  author={Liu, Baoyin and Yue, Ming and Liu, Rong},
  booktitle={2012 IEEE International Conference on Mechatronics and Automation},
  pages={1804--1809},
  year={2012},
  organization={IEEE}
}

@inproceedings{liu2024monorollbot,
  title={MonoRollBot: 3-DOF Spherical Robot with Underactuated Single Compliant Actuator Design},
  author={Liu, Zhiwei and Tafrishi, Seyed Amir},
  booktitle={IEEE International Conference on Soft Robotics (RoboSoft)},
  year={2025}
}

@article{tafrishi2019design,
  title={Design, modeling, and motion analysis of a novel fluid actuated spherical rolling robot},
  author={Tafrishi, Seyed Amir and Svinin, Mikhail and Esmaeilzadeh, Esmaeil and Yamamoto, Motoji},
  journal={Journal of Mechanisms and Robotics},
  volume={11},
  number={4},
  pages={041010},
  year={2019},
  publisher={American Society of Mechanical Engineers}
}

@article{xu2024tracked,
  title={Tracked robot with underactuated tension-driven RRP transformable mechanism: ideas and design},
  author={Xu, Ran and Liu, Chao},
  journal={Frontiers of Mechanical Engineering},
  volume={19},
  number={1},
  pages={4},
  year={2024},
  publisher={Springer}
}

@article{bons2023energy,
  title={An Energy-Dense Two-Part Torsion Spring Architecture and Design Tool},
  author={Bons, Zachary and Thomas, Gray C and Mooney, Luke and Rouse, Elliott J},
  journal={IEEE/ASME Transactions on Mechatronics},
  volume={29},
  number={3},
  pages={2373--2384},
  year={2023},
  publisher={IEEE}
}

@article{he2019underactuated,
  title={Underactuated robotics: a review},
  author={He, Bin and Wang, Shuai and Liu, Yongjia},
  journal={International Journal of Advanced Robotic Systems},
  volume={16},
  number={4},
  pages={1729881419862164},
  year={2019},
  publisher={SAGE Publications Sage UK: London, England}
}

@ARTICLE{10386077,

  author={Tafrishi, Seyed Amir and Hirata, Yasuhisa},

  journal={IEEE/ASME Transactions on Mechatronics}, 

  title={SPIRO: A Compliant Spiral Spring–Damper Joint Actuator With Energy-Based Sliding-Mode Controller}, 

  year={2024},

  volume={29},

  number={2},

  pages={947-959}}

@article{manti2016stiffening,
  title={Stiffening in soft robotics: A review of the state of the art},
  author={Manti, Mariangela and Cacucciolo, Vito and Cianchetti, Matteo},
  journal={IEEE Robotics \& Automation Magazine},
  volume={23},
  number={3},
  pages={93--106},
  year={2016},
  publisher={IEEE}
}

@article{liu2020survey,
  title={A survey on underactuated robotic systems: bio-inspiration, trajectory planning and control},
  author={Liu, Pengcheng and Huda, M Nazmul and Sun, Li and Yu, Hongnian},
  journal={Mechatronics},
  volume={72},
  pages={102443},
  year={2020},
  publisher={Elsevier}
}

@article{zhong2021toward,
	title={Toward Safe Human--Robot Interaction: A Fast-Response Admittance Control Method for Series Elastic Actuator},
	author={Zhong, Haoran and Li, Xinyu and Gao, Liang and Li, Congbo},
	journal={IEEE Transactions on Automation Science and Engineering},
	volume={19},
	number={2},
	pages={919--932},
	year={2021},
	publisher={IEEE}
}

@article{howard2013transferring,
	title={Transferring human impedance behavior to heterogeneous variable impedance actuators},
	author={Howard, Matthew and Braun, David J and Vijayakumar, Sethu},
	journal={IEEE Transactions on Robotics},
	volume={29},
	number={4},
	pages={847--862},
	year={2013},
	publisher={IEEE}
}

@article{vanderborght2013variable,
	title={Variable impedance actuators: A review},
	author={Vanderborght, Bram and Albu-Sch{\"a}ffer, Alin and Bicchi, Antonio and Burdet, Etienne and Caldwell, Darwin G and Carloni, Raffaella and Catalano, Manuel and Eiberger, Oliver and Friedl, Werner and Ganesh, Ganesh and others},
	journal={Robotics and autonomous systems},
	volume={61},
	number={12},
	pages={1601--1614},
	year={2013},
	publisher={Elsevier}
}

@article{mathews2022design,
	title={Design of Parallel Variable Stiffness Actuators},
	author={Mathews, Chase W and Braun, David J},
	journal={IEEE Transactions on Robotics},
	year={2022},
	publisher={IEEE}
}

@article{wolf2015variable,
	title={Variable stiffness actuators: Review on design and components},
	author={Wolf, Sebastian and Grioli, Giorgio and Eiberger, Oliver and Friedl, Werner and Grebenstein, Markus and H{\"o}ppner, Hannes and Burdet, Etienne and Caldwell, Darwin G and Carloni, Raffaella and Catalano, Manuel G and others},
	journal={IEEE/ASME transactions on mechatronics},
	volume={21},
	number={5},
	pages={2418--2430},
	year={2015},
	publisher={IEEE}
}

@article{kim2013compliant,
	title={Compliant joint actuator with dual spiral springs},
	author={Kim, Yongtae and Lee, Jimin and Park, Jaeheung},
	journal={IEEE/ASME Transactions on Mechatronics},
	volume={18},
	number={6},
	pages={1839--1844},
	year={2013},
	publisher={IEEE}
}

@article{wang2024rollbot,
  author    = {Wang, J. and Rubenstein, M.},
  title     = {Rollbot: a Spherical Robot Driven by a Single Actuator},
  journal   = {arXiv preprint arXiv:2404.05120},
  year      = {2024},
  doi       = {10.48550/arXiv.2404.05120}
}

@article{granularrobotics,
  author    = {Li, Han and others},
  title     = {Beyond Jamming Grippers: Granular Material in Robotics},
  journal   = {Advanced Robotics},
  volume    = {38},
  number    = {11},
  pages     = {715--729},
  year      = {2024},
  doi       = {10.1080/01691864.2024.2348544},
  url       = {https://doi.org/10.1080/01691864.2024.2348544}
}

\end{document}